\documentclass[letterpaper, 10 pt, conference]{ieeeconf}  
\AtBeginDocument{\let\autocite\cite}

\IEEEoverridecommandlockouts                              

\usepackage{balance}
\usepackage{url}
\usepackage{cite}
\usepackage{graphicx} 
\usepackage{flushend}
\usepackage{booktabs}
\usepackage{tikz}
\usepackage{amsmath}

\begin{document}
\IEEEoverridecommandlockouts
\overrideIEEEmargins

\title{\LARGE \bf
  Synthetic Emotions vs. Gamification: Exploring Engagement Strategies for Small Social Robots in Different Age Groups
}

\author{Morten Roed Frederiksen$^{1}$, and Kasper Stoy$^{2}$
  \thanks{{$^{1}$Morten Roed Frederiksen {\tt\small mrof@itu.dk} and $^{2}$Kasper Stoy \tt\small ksty@itu.dk} are affiliated with the Data Systems \& Robotics Department of The IT-University of Copenhagen.}
}

\maketitle

\begin{abstract}
Many children experience challenges in emotional regulation and social interaction, which can limit their participation in everyday activities and therapeutic programs. For socially assistive robots to be effective in this context, it is essential that children remain consistently and meaningfully engaged. We explore engagement strategies for a tactile robot designed to support children suffering from anxiety through daily interactions. The robot delivers either synthetic emotional feedback or point rewards to encourage user participation. We evaluated these strategies through two studies: a preference assessment with 16 school children aged 6–8 years, and a behavioral study with 14 university students aged 20–27 years in naturalistic environments. The study with school children indicated a preference for emotional engagement over points-based approaches. The follow up study with university students across a full day of interactions revealed contrasting results: points-based systems produced significantly higher task accuracy ($p < 0.05$) and sustained performance over time. Findings from different user groups suggest that stated preferences and behavioral outcomes can diverge depending on engagement context, highlighting the importance of validating design assumptions through observed interaction. This work contributes insights into age-related differences in engagement strategy effectiveness in human-robot interaction design.
\end{abstract}
\section{Introduction}
User engagement represents a critical design consideration in developing therapeutic social robots for children with mental health conditions. As interventions often rely on consistent daily interactions to achieve meaningful outcomes, ensuring sustained child engagement with robotic systems can become vital to the success of any intervention strategy \cite{Donnermann2024MeetOC, Kral2023SlowerRR, Balban2023BriefSR}. Previous research has demonstrated promising approaches to fostering engagement with social robots. These strategies encompass adaptive behaviors that respond to individual user needs \cite{Shenoy2022ASL, Bajones2016EnablingLH}, multi-modal interactions that leverage multiple sensory channels \cite{Maniscalco2022BidirectionalMS, Lima2021RoboticTF, Applewhite2021NovelBM}, enhanced predictability to reduce user anxiety \cite{Schadenberg2021PredictableRF}, and affective narratives that create emotional connections \cite{Liang2023ARD, Frederiksen2022RobotVA, Park2022EmpathyIH}. Most implementations either dynamically adapt robot behaviors to optimize user-specific scenarios \cite{Burns2023ALI, Weber2018HowTS} or focus on entertainment-driven interactions \cite{Anhuaman2023CoguiIS, Ahmad2020VideoLT, Tisza2024CanRE}.

Robot engagement involves multiple design factors\cite{Sorrentino2024FromTD}, including gaze patterns \cite{Somashekarappa2024GoodLH}, morphological changes \cite{Vlachos2016TheEO,Yim2024IntegrationOA}, behaviors, and narratives \cite{Spitale2022SociallyAR,Frederiksen2022RobotVA}. Paralleling these developments, computer gaming research has extensively explored user motivation through engagement strategies including narrative design \cite{McClintock2024AnalyzingDI}, usability optimization \cite{lvarezXochihua2017ComparingUU}, gameplay mechanics \cite{Dickey2005EngagingBD}, and character perception \cite{Lankoski2011PlayerCE}. These studies reveal common patterns in how users are motivated to interact with technology, suggesting potential crossover applications for social robotics.
\begin{figure}[h]
\centering
\includegraphics[width=0.48\textwidth]{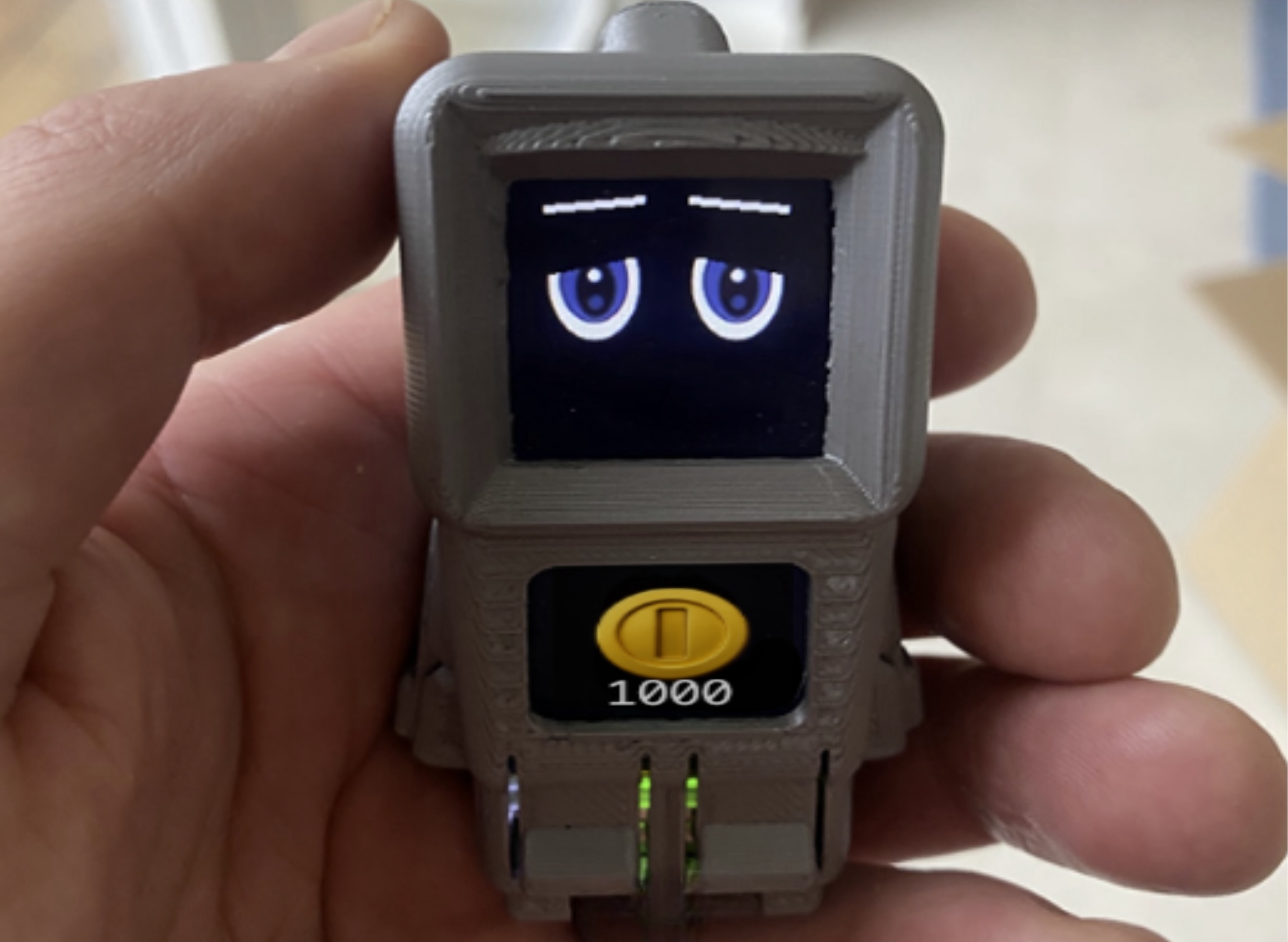}
\caption{The AffectaPocket robot used in the experiments. The robot here is depicted with the points system enabled. Each time a participant would interact with the robot and complete a small tactile interaction, they would be rewarded with a coin animation and 1000 points.}
\label{affecta_front_page}
\end{figure}
This paper explores the design and evaluation of two distinct engagement strategies for children interacting with a pocket-sized tactile robot developed to aid those suffering from anxiety disorders in the long term. The first strategy leverages synthetic emotions to create empathetic connections, while the second employs a gamification \cite{Deterding2011FromGD} approach inspired by traditional reward systems, specifically implementing a Super Mario-inspired coin and points-based mechanism \cite{Schneider2017AnalysisOP,Rutledge2018GamificationIA}.

We evaluated these engagement strategies through two user studies. A preference assessment study with 16 school children aged 6-8, which indicated user preference for emotional engagement strategies over points-based approaches (68.75\% preference rate). We followed this initial study up with a behavioral study involving 14 university students in naturalistic settings, measuring actual engagement behaviors rather than stated preferences over a full day of interactions. In the following sections, we describe the two studies designed to compare emotional and gamified engagement strategies and examine how user age and interaction context shape engagement outcomes


\section{Related Work}

\subsubsection{Therapeutic Robots and Engagement Strategies}

Paro the seal robot exemplifies how robots have used morphological aspects to entice humans to interact with the robot \cite{Shoesmith2024UsingPA,Inoue2021ExploringTA}. The robot has a soft fur outer surface that feels comfortable to hold and to stroke, mimicking the behavior of a lapdog. Other studies have tried to strengthen the anthropomorphic interpretation of robots by letting children take care of the robots' artificial 'needs' \cite{Cagiltay2022ExploringCP, Hieida02012022}.

Much of the previous research on engagement in Human-Robot Interaction focuses on estimating the human level of engagement during an interaction either based on social cues \cite{Ghazali2017PardonTR, Ravandi2025DeepLA}, tone of voice \cite{Oertel2020EngagementIH, Schuller2011RecognisingRE}, gaze patterns \cite{Fiore2013TowardUS}, or semantic content of the interaction \cite{Maniscalco2022BidirectionalMS, DelDuchetto2020AreYS}. However, fewer studies have focused on designing specific strategies to enhance engagement, particularly comparing fundamentally different approaches.

Research has explored two primary engagement paradigms: emotional connection and structured rewards. Emotional approaches leverage synthetic emotions to create empathetic connections \cite{Park2022EmpathyIH}, with studies showing that robots displaying vulnerability can elicit user empathy and stronger emotional bonds \cite{Frederiksen2022RobotVA}. Adaptive emotional systems that respond to user states have demonstrated sustained engagement over time \cite{Shenoy2022ASL, Burns2023ALI}. Conversely, gamification approaches employ extrinsic motivational mechanisms such as points, badges, and achievement systems to drive user behavior \cite{Deterding2011FromGD}. While initially effective, longitudinal studies suggest that simple reward mechanisms may face diminishing returns over extended periods without proper adaptation \cite{Khne2024HowDC}.

Age-related differences in robot engagement reveal important considerations for therapeutic applications. Children show greater tendency to anthropomorphize robots and engage with emotional expressions \cite{Beer2015YoungerAO}, while predictable robot behavior is crucial for maintaining engagement in vulnerable populations such as children with anxiety disorders \cite{Schadenberg2021PredictableRF}. Previous research results have also suggested that the anticipation of engagement with a robot could help lessen some of the immediate anxiety symptoms in high-anxiety patients \cite{Nomura2019DoPW}.

Despite extensive research on individual engagement strategies, few studies have directly compared the effectiveness of emotional versus reward-based approaches across different user populations and interaction contexts. Additionally, limited research has examined potential disconnects between user preferences and actual behavioral engagement, particularly in therapeutic applications where sustained long-term interaction is crucial for treatment success.

\section{Method and study design}
\subsection{A pocket-sized tactile robot}
\begin{figure}[h]
\centering
\includegraphics[width=0.48\textwidth]{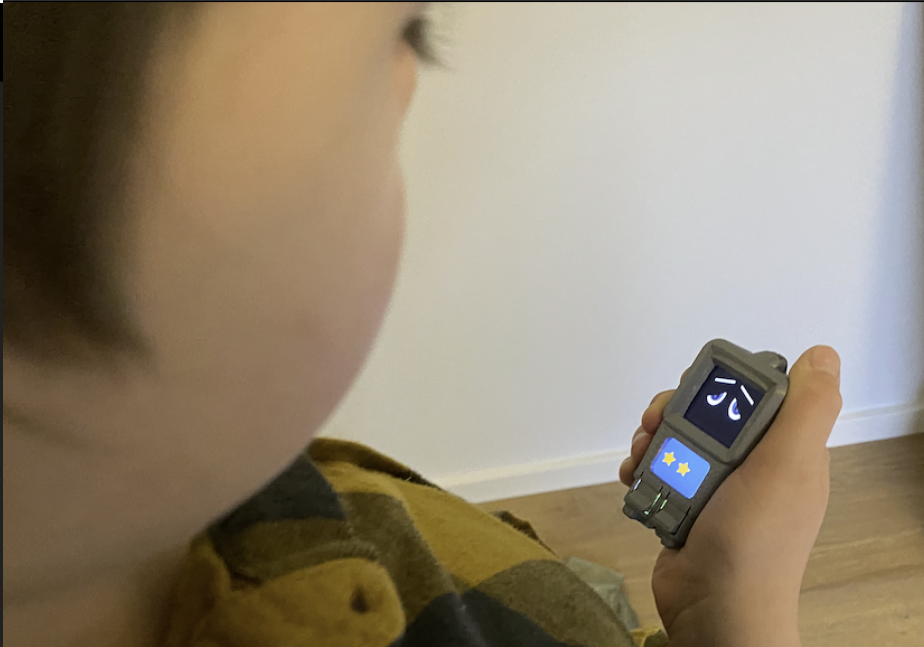}
\caption{A child interacts with the robot with the synthetic emotion system enabled. The stars on the torso screen of the robots depicts a tutorial process that the children would initial go through to learn the basic concepts about the tactile interaction. Once learned the robot would no longer display this visual aid on its torso screen.}
\label{affecta_synthetic}
\end{figure}

We used an AffectaPocket robot originally developed by these authors for therapeutic applications with children \cite{Frederiksen2024TactileCL}. The robot can be seen in Figure \ref{affecta_front_page}, and Figure \ref{affecta_synthetic}. It is a pocket-sized socially assistive robot developed to support children with anxiety through discrete, tactile interaction. It is designed to be grasped and used in the child’s pocket, enabling private interaction in socially sensitive settings. The robot features a 3D-printed outer shell with anthropomorphic elements, a torso-mounted screen capable of displaying simple facial expressions, and five capacitive touch sensors—two on the sides, one on the front, and two at the back—to detect grasping gestures. Interaction with the robot is centered on a silent rhythm-matching game: when the robot is grasped, it vibrates in a three-note pattern that the user must replicate by timing their squeezes. In training mode, these rhythms are also visualized as stars on the torso screen to help children form an initial mental model of the task. Once the child has learned the interaction, the robot can be used entirely through tactile input without the need for visual or auditory cues. The robot is controlled by an ESP32 chip and powered by a rechargeable battery. The AffectaPocket’s facial display does not use a fixed set of discrete expressions but instead implements a continuous, stepless emotion mapping. The robot dynamically adjusts the angle and curvature of the eyes and eyebrows according to the calculated mood value (as can be seen in Figure \ref{affecta_synthetic}). This allows for smooth transitions between affective states from ‘very sad’ to ‘very happy.’ The expressions are rendered on a 1.9 inch LCD screen, enabling subtle visual feedback that reflects gradual emotional changes rather than abrupt categorical shifts.



\subsection{Study 1: Preference Assessment: Children aged 6-8.}
Following the pilot findings, we conducted a survey study with school children (N=16) to investigate preferences between different robot engagement strategies. The survey asked the children to compare two distinct approaches: (1) a points-based reward system and (2) an emotions-based system where participants could influence the robot's synthetic emotional states. Participants were demonstrated both approaches and then completed a structured questionnaire assessing explicit strategy preferences using forced-choice questions and rating their anticipated motivation for each approach on 10-point scales (1=not motivating at all, 10=extremely motivating). Each interaction session lasted approximately 30 minutes and No monetary rewards or grades were provided. Participants only received the robot’s feedback (emotional or point-based) as incentive. We calculated frequency distributions and percentages for categorical preference responses. Motivation ratings between the two conditions were compared using paired samples t-tests, with Cohen's d calculated for effect size estimation. All analyses were conducted with alpha set at .05.

\subsection{Study 2: Behavioral study: University students in Natural Setting}
We conducted a between-subjects field study (n=14) comparing actual engagement behaviors between emotions-based and points-based robot systems. Fourteen college students were randomly assigned to interact with either an emotions-based robot (n=7) or points-based robot (n=7) for one full day in their natural environments. The system logged all interaction events including game attempts, physical touches, and task performance metrics. Primary outcome measures included task accuracy (percentage of correctly matched items per game), total engagement volume (sum of all game attempts and physical interactions), and temporal engagement patterns (engagement per time spent with the robot). Each interaction session lasted approximately 6 hours. Following the interaction period, participants completed a post-interaction questionnaire assessing their subjective experience with the robot. The questionnaire included items measuring overall enjoyment, perceived robot responsiveness, emotional connection, and likelihood of future interaction, using 7-point Likert scales. We also collected qualitative feedback about their experience through open-ended questions. One of the datasets were missing a lot of entries for one of the with-interaction condition. Any missing data points were added via mean imputation. Mean imputation was chosen due to the low proportion of missing data and the relatively small sample size, where more complex methods (e.g., multiple imputation) might have introduced additional variance without meaningful benefit. All imputations were performed separately per condition to preserve within-group distributions. We analyzed between-group differences using Mann-Whitney U tests for engagement volume metrics (total game attempts and physical interactions) due to non-normal distribution confirmed by Shapiro-Wilk tests (p < .05). For task accuracy data, which met normality assumptions, we used independent samples t-tests. Effect sizes were calculated using ranked biserial correlation (r) for Mann-Whitney U tests and Cohen's d for t-tests. Post-interaction questionnaire responses (7-point Likert scales) were analyzed using independent samples t-tests with Cohen's d effect sizes.


\subsection{Synthetic emotions}

The synthetic emotional system was designed to move beyond simple linear responses that react immediately to input and decay at a fixed rate. Such systems may risk encouraging brief bursts of interaction rather than supporting more sustained engagement. To address this, the model incorporates a gradually changing mood state influenced by both the timing and regularity of user interactions. Drawing inspiration from homeostatic systems like insulin-glucose regulation, it uses a decay mechanism and an adaptive impact function where closely spaced interactions contribute less to emotional change. This allows the robot’s affective state to evolve over time, making it more sensitive to consistent interaction patterns than to high-frequency input. The intention is not to replicate human emotion, but to create a dynamic that may encourage ongoing engagement and make the robot’s responses feel more socially meaningful over time.
The robot maintains a mood variable $M(t) \in [1,100]$, which evolves over time based on user interactions and autonomous decay. In the absence of stimulation, mood passively decays at discrete 1-second intervals with an adaptive decay rate $\delta(t)$, which slowly increases over time to simulate metabolic resistance analogous to insulin resistance. Specifically, mood decay is updated every 1000 milliseconds as:
$$M(t) =  M(t-\Delta t) - \delta(t) \cdot \Delta t$$
$$\delta(t) = \delta(t-\Delta t) + \frac{1}{10^7}$$
where $\Delta t$ is the elapsed time since the last decay step in milliseconds. we constrain mood M(t) to the range [1,100], and decay
$\delta(t)$ to a maximum of 0.0001. These bounds are meant to prevent runaway growth or vanishing values. When a positive interaction occurs (e.g., task completion), the system increases mood by an amount dependent on an impact variable $I(t)$. The impact increases linearly with time since the last interaction, but only for the first hour (3600 seconds), after which it plateaus:
$$I(t) = \begin{cases}
I(t_0) + 0.75 \tau & \text{if } \tau < 3600\text{s} \\
I(t_0) + 2700 & \text{if } \tau \geq 3600\text{s}
\end{cases}$$
where $\tau$ is the time elapsed since the last interaction in seconds, and $I(t_0)$ is the impact value at the time of the last interaction. Upon each interaction, the mood and impact are updated as:
$$M(t) = M(t) + I(t) \cdot 50, \quad \delta(t) = \delta(t) \cdot 0.0005$$
$$I(t) = \max(0, I(t) \cdot 0.75)$$
These equations ensure diminishing returns on frequent interactions (as $I(t)$ decays rapidly after use via the 0.75 multiplicative factor), increasing emotional sensitivity after prolonged inactivity. The calculated mood was directly tied to the facial expression of the robot as seen in Figure \ref{affecta_synthetic}, that used its eye and eyebrows to convey sadness for lower range moods and with a high mood of 100 expressed happiness. This formulation was developed by the authors, inspired by homeostatic regulation models in biological systems (e.g., insulin-glucose feedback). It was not adopted from existing emotion models, but derived empirically through iterative testing to produce stable mood dynamics in the AffectaPocket platform.

\begin{figure*}[ht]
    \centering
    \includegraphics[width=\textwidth]{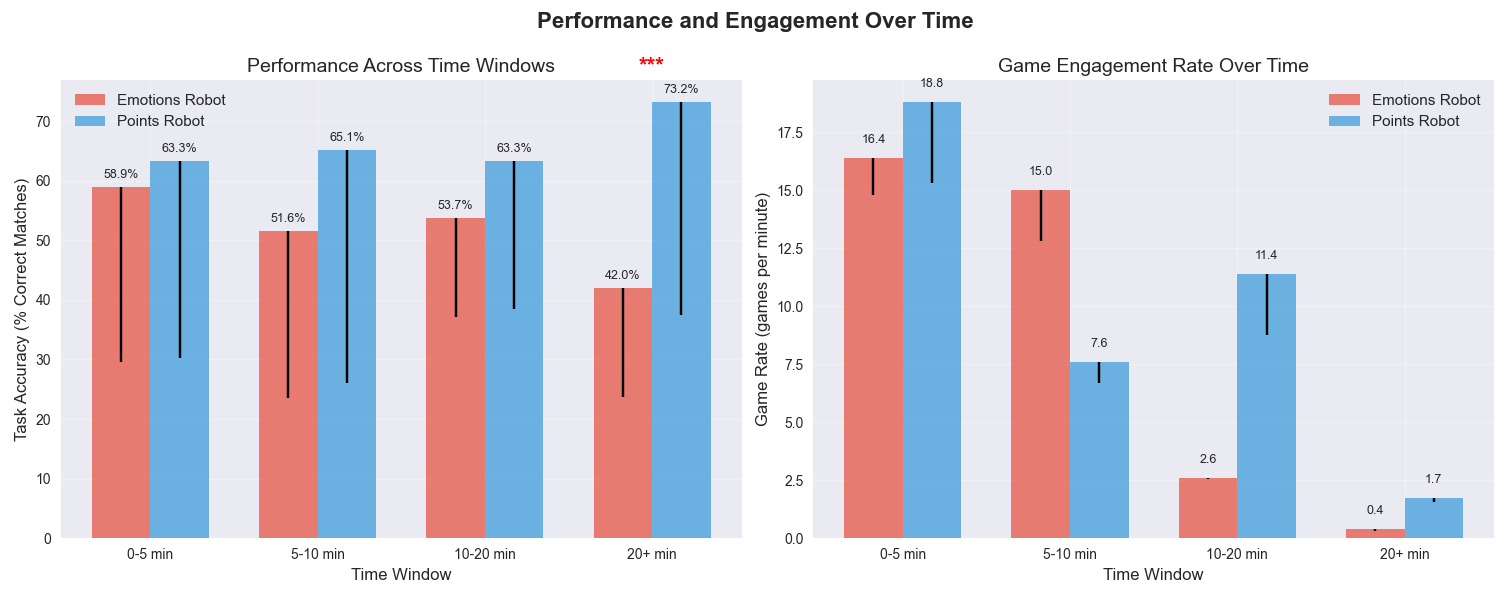}
    \caption{Temporal analysis of robot engagement strategies for the second study. (A) Task accuracy across four time windows showing declining performance for emotions-based robots (red bars) versus sustained performance for points-based robots (blue bars). (B) Game engagement rates (denoted as average number of games played per minute) demonstrating consistent interaction patterns over time. Statistical significance markers (***) indicate p $<$ 0.05. Error bars represent standard error of the mean. Points-based systems produced significantly higher task accuracy in later time windows (10-20 min and 20+ min) while maintaining stable engagement rates throughout the session.}
    \label{engagement_findings}
\end{figure*}
\subsection{Gamification Rewards}
As a control condition, we implemented a traditional gamification system based on extrinsic reward mechanisms commonly found in digital games and educational applications \cite{Deterding2011FromGD, McGonigal2011RealityIB}. Following established principles from play theory, this system employed immediate positive reinforcement through point accumulation to drive user engagement \cite{Zhi2024GamefulIH}. Upon completion of each interaction task, users received exactly 1000 points, displayed through an animated spinning coin symbol on the robot's torso screen as seen in Figure \ref{affecta_front_page}. This visual feedback mechanism draws from operant conditioning principles where consistent reward schedules maintain behavioral patterns [5]. To isolate the effects of the reward system from emotional variables, the robot's internal mood was fixed at the maximum value of 100 throughout all interactions, with the robot's eye display maintaining a consistently happy expression. This design choice prevented any confounding effects between the gamification rewards and synthetic emotional responses, ensuring that user engagement differences could be attributed solely to the contrasting motivational frameworks rather than mixed emotional and reward signals [6,7].

\section{Results}
We used stated preference as the primary measure for children due to ethical and logistical constraints on prolonged behavioral testing with minors. In contrast, university participants engaged in unsupervised, full-day interactions, enabling continuous behavioral logging. This methodological difference explains why preference was measured for children and engagement for adults.

\subsection{Study 1: The Preference Assessment}
Survey data revealed strong user preference for emotionally-based engagement strategies over traditional gamification approaches. A clear majority (68.8\%) explicitly stated preference for making the robot happy over accumulating points. This preference pattern was supported by self-reported motivation ratings, with participants reporting higher anticipated engagement for emotional strategies (M=8.7 out of 10, SD=1.2) compared to points-based approaches (M=8.00 out of 10, SD=1.4), though this difference approached but did not reach statistical significance (t(15)=1.97, p=.07). Participants demonstrated high confidence in their perceived ability to engage with the robot's emotional states, rating their capacity to help a "sad" robot at M=8.7 out of 10 (SD=1.3). These findings highlight the appeal of emotionally expressive systems among school-aged children and provided the basis for exploring how different engagement strategies affect actual interaction behavior in a separate, context-specific study.

\subsection{Study 2: The Behavioral Exploration}
Despite stated preferences favoring emotional engagement, behavioral data demonstrated substantial performance advantages for points-based systems across multiple metrics as depicted in Figure \ref{engagement_findings}. Most notably, participants in the points condition achieved significantly higher task accuracy (M=69.4\%, SD=13.2) compared to the emotions condition (M=47.7\%, SD=7.1), U=6, p=.02, r=.7 showing a large effect size. This performance gap became more pronounced during extended interactions, with points-based participants maintaining 73.2\% accuracy after 20 minutes compared to 41.7\% for emotions-based participants (t(12)=-8.5, p$<$.001, d=-0.97).

Engagement volume metrics (number of games played per time unit) showed consistent trends favoring the points condition, though not all reached statistical significance. Points-based participants attempted 105 games on average (SD=90.5) compared to 46 (SD=32) in the emotions condition, representing a large effect size (d=-.87) despite non-significance (p=.13). Analysis of temporal patterns revealed divergent trajectories: emotions-based participants showed declining performance from early (55.7\%) to late sessions (41.5\%), while points-based participants improved from 65.1\% to 73.6\% accuracy over time. These behavioral patterns suggest that while emotional engagement strategies may be preferred conceptually, points-based systems more effectively sustain performance and engagement in practice.

As for the post-interaction questionnaire, the Points-based participants rated their robot as more humanlike (M=3.1, SD=0.2 vs M=2.5, SD=0.7) and likable (M=4.1, SD=0.6 vs M=3.6, SD=0.7), though these large effect sizes did not reach statistical significance with the current sample size. Notably, this suggest an opposite effect than found in common assumptions that emotional expressiveness is necessary for positive robot perception. Within-condition motivation ratings were moderate for both mood-based (M=5.1, SD=3.1) and points-based (M=6.6, SD=2.6) feedback, indicating that neither approach was overwhelmingly compelling, despite the indicated behavioral performance advantage of the points system. The standard deviations also reveals an interesting pattern: the points condition showed much more consistent ratings for human-like qualities (SD=0.2 vs 0.7), suggesting more uniform perception, while the emotions condition showed high variability in motivation ratings (SD=3.1), indicating that emotional feedback worked very well for some participants but poorly for others.

\section{Discussion}

\subsection{Developmental and Motivational Alignment with Engagement Strategies}
Although the study did not reveal a statistically significant overall preference for either engagement modality, there were observable tendencies that may reflect underlying developmental or motivational differences across age groups. These findings should be interpreted with caution given the small sample sizes in both studies. Larger-scale replications are needed to confirm the observed trends. However, the participants at age 6-8 appeared to respond more positively to synthetic emotional expressions, potentially due to a heightened tendency to anthropomorphize robots and to interpret affective cues as socially meaningful \cite{Beer2015YoungerAO, Berghe2019ATO}. This aligns with prior findings suggesting that children often engage with social robots in ways that emphasize emotional reciprocity and empathetic responses \cite{Manzi2020ARI,Khne2024HowDC}. In contrast, the study on older participants, showed that the university students, seemed to favor the point-based system, which may correspond with a preference for structured, goal-oriented interaction models and the influence of extrinsic motivational factors \cite{Helln2023EnhancingSM, Riedmann2022IntegrationOA}. However, these patterns must be interpreted with caution, as factors such as task framing, social context, and interaction expectations are known to influence user engagement and may have impacted any age-based effects \cite{OBrien2018APA,Nwaimo2024DatadrivenSF}. It is also important to note that the AffectaPocket was originally designed for children aged 6–8, and not optimized for adult users. The smaller form factor and visual design may have reduced perceived relevance among university students. 
Overall these findings suggest that engagement strategies in social robots could potentially benefit from approaches that are sensitive to both developmental characteristics and contextual variables, for instance through adaptive systems that can tailor interaction modalities to individual user profiles and situational demands. We suggest conducting behavioral validation studies specifically with children in the target age range to determine whether the performance advantages of points-based systems observed in university students translate to pediatric populations, or whether developmental differences in motivation and cognition lead to different engagement patterns.

\subsection{Temporal Considerations and the Importance of Longitudinal Assessment}
Our study focus on short-term interaction outcomes, which may not adequately capture the long-term effectiveness or sustainability of different engagement strategies. Initial user preferences often reflect novelty effects or surface-level appeal, and may not correspond to enduring engagement patterns \cite{Khne2024HowDC,Paschmann2024DrivingMA}. For instance, while synthetic emotional responses may elicit an empathetic reaction during brief interactions, their impact may diminish over time if users begin to perceive them as inauthentic or repetitive, particularly in older age groups \cite{RosenthalvonderPtten2019NeuralMF}. Also point-based systems may offer initial clarity and structure, but risk reduced engagement if the rewards are not sufficiently meaningful or fail to evolve with the interaction \cite{hanus2015, Khne2024HowDC,Paschmann2024DrivingMA}. These concerns highlight the need for longitudinal studies to investigate how engagement strategies perform over extended periods and to evaluate whether hybrid approaches, that combine intrinsic (emotional) and extrinsic (points-based) motivation, can offer more robust and context-sensitive engagement frameworks. For example, points could unlock emotional responses or vice versa, encouraging balanced engagement dynamics across user groups.

\section{Conclusion}

This paper presents preliminary findings from a set of studies investigating engagement strategies for a pocket-sized social robot designed to support therapeutic interaction. We implemented and evaluated two interaction models: one based on bio-inspired synthetic emotions and another using gamified point rewards. The studies explored how these strategies affected engagement in different user segments and contexts. One key observation was that a gamified reward system led to more sustained interaction in a naturalistic setting with older users.

These results offer early insight into how interaction design choices shape user engagement and suggest directions for developing adaptive systems tailored to different use cases. Future work should focus on longitudinal deployment with the target therapeutic user group to examine how engagement strategies influence outcomes over time.

\balance

\bibliography{bibliography}
\bibliographystyle{IEEEtran}

\end{document}